%% file: main.tex
\documentclass[11pt,a4paper]{article}

\usepackage{amsmath}
\usepackage{amssymb}
\usepackage{amsfonts}
\PassOptionsToPackage{hyphens}{url}
\usepackage{hyperref}
\usepackage{color}
\usepackage{authblk}
\usepackage{graphicx}
\usepackage{subcaption}
\graphicspath{ {./figures/} }

\newcommand{\code}[1]{\texttt{#1}}

\begin{document}
\title{Descriptive and Predictive Analysis of Euroleague Basketball Games and the Wisdom of Basketball Crowds}
\author{Georgios Giasemidis}
\date{}

\affil{g.giasemidis@gmail.com, Reading, UK}

\maketitle

\input{abstract}

\input{introduction}

\input{data}

\input{analysis}

\input{modelling}

\input{conclusions}

\bibliographystyle{plain}
\bibliography{bib}

\end{document}

%% file: abstract.tex

\begin{abstract}
	In this study we focus on the prediction of basketball games in the Euroleague competition using machine learning modelling. The prediction is a binary classification problem, predicting whether a match finishes 1 (home win) or 2 (away win). Data is collected from the Euroleague's official website for the seasons 2016--2017, 2017--2018 and 2018--2019, i.e. in the new format era. Features are extracted from matches' data and off-the-shelf supervised machine learning techniques are applied. We calibrate and validate our models. We find that simple machine learning models give accuracy not greater than 67\% on the test set,  worse than some sophisticated benchmark models. Additionally, the importance of this study lies in the ``wisdom of the basketball crowd'' and we demonstrate how the predicting power of a collective group of basketball enthusiasts can outperform machine learning models discussed in this study. We argue why the accuracy level of this group of ``experts'' should be set as the benchmark for future studies in the prediction of (European) basketball games using machine learning.
\end{abstract}

%% file: introduction.tex

\section{Introduction}
\label{sec:intro}

Basketball is changing with fast pace. It is broadly considered that the game has entered a new era, with a different style of play dominated by three-point shots. An enjoyable and illustrative overview of the main changes is given in \cite{sprawnball}, where the author demonstrates the drastic changes in the game through graph illustrations of basketball statistics.

Advanced statistics have played a key-role in this development. Beyond the common statistics, such as shoot percentages, blocks, steals, etc., the introduction of advanced metrics, (such as the ``Player Efficiency Rate'', ``True Percentage Rate'', ``Usage Percentage'', ``Pace factor'', and many more), attempt to quantify ``hidden" layers of the game, and measure a player's or a team's performance and efficiency on a wide range of fields. In Layman's terms, the following exampel is illustrative. A good offensive player, but bad defender, might score 20 points on average, but concede 15. An average offensive player, but good defender might score on average 10 points, but deny 10 points (i.e. 5 good defensive plays), not through blocks or steals, but through effective personal and team defense. Using traditional basketball statistics, the former player scores more points but overall is less efficient to the team. 

A major contribution to the field of advanced basketball statistics is made in \cite{Oliver20014}, where the author ``\textit{demonstrates how to interpret player and team performance,  highlights general strategies for teams when they're winning or losing and what aspects should be the focus in either situation.}''. In this book, several advanced metrics are defined and explained. In this direction, further studies have contributed too, see \cite{Kubatko2007} and \cite{RustThesis2014}.

With the assistance of the advanced statistics modern NBA teams have optimised the franchise's performance, from acquiring players with great potential, which is revealed though the advanced metrics, a strategy well-known as \textit{money-ball}, to changing the style of the game. A modern example is the increase in the three-point shooting, a playing strategy also known as ``\textit{three is greater than two}''.

Despite the huge success of the advanced statistics, these measures are mostly descriptive. Predictive models started arising for NBA and college basketball in the past ten years.

The authors in \cite{Beckler2009nbao} first used machine learning algorithms to predict the outcome of NBA basketball games. Their dataset spans the seasons 1991-1992 to 1996-1997. They use standard accumulated team features (e.g. rebounds, steals, field goals, free throws etc.), including offensive and defensive statistics together with wins and loss, reaching 30 features in total. Four classifiers were used, linear regression, logistic regression, SVMs and neural networks. Linear regression achieves the best accuracy, which does not exceed 70\% on average (max 73\%) across seasons. The authors observe that accuracy varies significantly across seasons, with ``\textit{some seasons appear to be inherently more difficult to predict}''. An interesting aspect of their study is use of the ``NBA experts'' prediction benchmark, which achieves 71\% accuracy.

In \cite{Torres2013pdn}, the author trains three machine learning models, linear regression, maximum likelihood classifier and a multilayer perceptron method, for predicting the outcome of NBA games. The dataset of the study spans the seasons 2003--2004 to 2012--2013. The algorithms achieve accuracy 67.89\%, 66,81\% and 68.44\% respectively. The author used a set of eight features and in some cases PCA boosted the performance of the algorithms.

The authors of \cite{zimmermann2013predicting} use machine learning approaches to predict the outcome of American college basketball games (NCAAB). These games are more challenging than NBA games for a number of reasons (less games per season, less competition between teams, etc.). They use a number of machine learning algorithms, such as decision trees, rule learners, artificial neural networks, Naive Bayes and Random Forest. Their features are the four factors and the adjusted efficiencies, see \cite{Oliver20014,zimmermann2013predicting}, all team-based features. Their data-set extends from season 2009 to 2013. The accuracy of their models varies with the seasons, with Naive Bayes and neural networks achieving the best performance which does not exceed 74\%. An interesting conclusion of their study is that there is \textit{``a “glass ceiling” of about 74\% predictive accuracy that cannot be exceeded by ML or statistical techniques''}, a point that we also discuss in this study.

In \cite{Fang2015ngp}, the authors train and validate a number of machine learning algorithms using NBA data from 2011 to 2015. They assess the performance of the models using 10-fold cross-validation. For each year, they report the best-performing model, achieving a maximum accuracy of 68.3\%. They also enrich the game-data with players' injury data, but with no significant accuracy improvements. However, a better approach for validating their model would be to assess it on a hold-out set of data, say the last year. In their study, different algorithms achieve the best performance in every season, which does not allow us to draw conclusions or apply it to a new dataset; how do we know which algorithm is right for the next year?

The authors of \cite{Jain2017mla} focused on two machine learning models, SVM and hybrid fuzzy-SVM (HFSVM) to predict NBA games. Their data-set is restricted to the regular season 2015-2016, which is split into training and test sets. Initially, they used 33 team-level features, varying from basic to more advanced team statistics. After applying a feature selection algorithm, they concluded to 21 most predictive attributes. The SVM and HFSVM achieve 86.21\% and 88.26\% average accuracy respectively on 5-fold cross-validation. Their high accuracy though could be an artefact of the restricted data-set, as it is applied to a single season, which could be an exceptionally good season for predictions. A more robust method to test their accuracy would be to use $k$-fold cross validation on the training set for parameter tuning and then validate the accuracy of the model on a hold-out set of games (which was not part of cross-validation). Otherwise, the model might be over-fitted. 

In \cite{Oursky2019mla}, the authors use deep neural networks to predict the margin of victory (MoV) and the winner of NBA games. They treat the MoV as a regression model, whereas the game winner is treated as a classification model. Both models are enhanced with a RNN with time-series of players' performance. Their dataset includes data from 1983. They report an accuracy of 80\%  on the last season.

Although advanced statistics have been introduced for more than a decade in the NBA, European basketball is only recently catching up. The Euroleague \footnote{The Euroleague is the top-tier European professional basketball club competition.} organised a student's competition in 2019 \footnote{The SACkathon, see \url{http://sackathon.euroleague.net/}.}, whose aim was ``\textit{to tell a story through the data visualization tools}''.

Despite the oustanding progress on the prediction of NBA games, European basketball has been ignored in such studies. European basketball has a different playing style, the duration of the games is shorter and even some defensive rules are different. Applying models trained on NBA games to European games cannot be justified. 

Predicting Euroleague games has a number of challenges. First, the Euroleague new format is only a few years old, therefore data is limited compared to the NBA, where one can use tens of seasons for modelling. Also, the number of games is much lower, 30 in the first three seasons compared to about 80 games in NBA. Therefore, the volume of data is much smaller. Secondly, Euroleague teams change every season. Euroleague has closed contracts with 11 franchises which participate every year, however the remaining teams are determined through other paths \footnote{\url{https://www.euroleaguebasketball.net/euroleague-basketball/news/i/6gt4utknkf9h8ryq/euroleague-basketball-a-licence-clubs-and-img-agree-on-10-year-joint-venture}}. Hence, Euroleague is not a closed championship like the NBA,  teams enter and leave the competition every year, making it more challenging to model the game results.

When the current study started, no study existed that predicted European basketball games. During this project, a first attempt was made by the Greek sport-site \url{sport24.gr}, which was giving the probabilities of the teams winning a Euroleague game, probabilities which were updated during the game. Their model was called the ``Pythagoras formula'' \footnote{\url{https://www.sport24.gr/Basket/live-prognwseis-nikhs-sthn-eyrwligka-apo-to-sport24-gr-kai-thn-pythagoras-formula.5610401.html}}. According to this source, ``\textit{the algorithm was developed by a team of scientists from the department of Statistics and Economics of the University of Athens in collaboration with basketball professionals. The algorithm is based on complex mathematical algorithms and a big volume of data}''. However, this feature is no longer available on the website. Also, no performance information is available about it.

More recent attempts on predicting European basketball games can be found on Twitter accounts, which specialise in basketball analytics. Particularly, the account ``cm simulations'' (\url{@cm_silulations}) gives the predictions of Euroleague and Eurocup games each week for the current season 2019-2020. On a recent tweet \footnote{\url{https://twitter.com/CmSimulations/status/1225901155743674368?s=20}, \url{https://twitter.com/CmSimulations/status/1226233097055850498?s=20}}, they claim that they have achieved an accuracy of 64.2\%, for rounds 7 to 24 of this season. However, no details of the algorithm, the analysis and the methodology have been made public.

In this article, we aim to fill this gap, using modern machine learning tools in order to predict \textit{European} basketball game outcomes. Also, we aim to underline the machine-learning framework and pipeline (from data collection, feature selection, hyper-parameter tuning and testing) for this task, as previous approaches might have ignored one or more of these steps. 

From the machine-learning point of view, this is a binary classification problem, i.e. an observation must be classified as 1 (home win) or 2 (away win). For our analysis we focus on the Euroleague championship in the modern format era covering three seasons, 2016-2017, 2017-2018 and 2018-2019. The reason we focus on the Euroleague is entirely personal, due to the experience and familiarity of the authors with the European basketball. The reason we focus on the latest three seasons is that prior to 2016, the championship consisted of several phases in which teams were split in several groups and not all teams played against each other. In the modern format, there is a single table, where all teams play against each other. This allows for more robust results.

We should emphasize that the aim of the paper is to extend and advance the knowledge of the machine learning and statistics in basketball, no-one should use it for betting purposes. 

This article is organised as follows. In Section \ref{sec:data}, the data-set, its size, features, etc., is described, whereas in Section \ref{sec:descriptive} a descriptive analysis of the data-set is presented. Section \ref{sec:predictive-modelling} focuses on the machine-learning predictive modelling and explains the calibration and validation processes, while it also discusses further insights such as the ``wisdom of the crowd''. We summarise and conclude in Section \ref{sec:conclusions}

%% file: data.tex

\section{Data Description}
\label{sec:data}

The data is collected from the Euroleague's official web-site \footnote{\url{https://www.euroleague.net/}} using scraping methods and tools, such as the \code{Beautiful Soup} package \footnote{\url{https://pypi.org/project/beautifulsoup4/}} in \code{Python}. The code for data extraction is available on Github \footnote{\url{https://github.com/giasemidis/basketball-data-analysis}}. The data collection data focuses on the regular seasons only.

\subsection{Data Extraction}

Data is collected and organised in three types, (i) game statistics, (ii) season results and (iii) season standings. 

Game statistics includes statistics of each team in a game, such as the offense, defense, field goal attempted and made, percentages, rebounds, assists, turnovers etc. Hence, each row in the data represents a single team in a match and its statistics.

Season results data files include game-level results, i.e. each row correspond to a game. The data file stores the round number, date, home and away teams and their scores.

Season standings data files contains the standings with the total points, number of wins and losses, total offense, defense and score difference at the end of each round.

%% file: analysis.tex

\section{Descriptive Analysis}
\label{sec:descriptive}

In this section, we perform a descriptive analysis of the data, understanding some key aspects of the European basketball in the new format era, i.e. the seasons 2016 - 2017, 2017 - 2018 and 2018 - 2019. For simplicity, we refer to each season using the end year of the season, e.g. 2016 - 2017 is referred to as season 2017 thereafter. 

\begin{figure}[]
	\centering
	\includegraphics[scale=0.5]{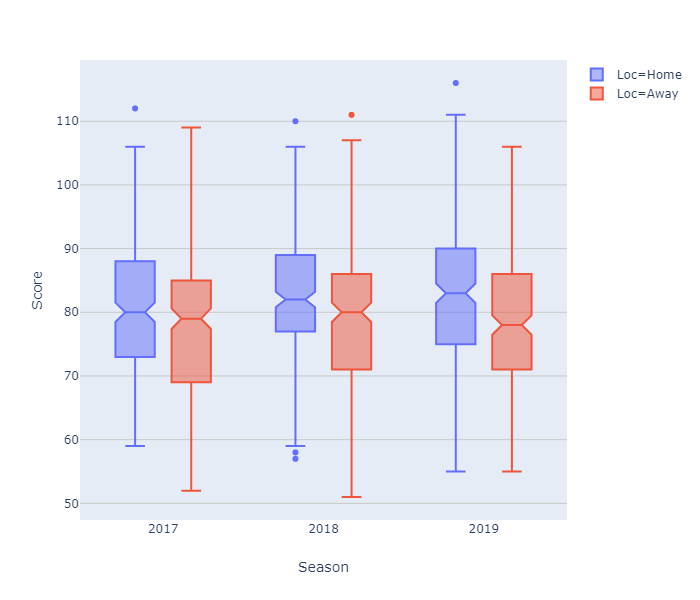}
	\caption{Box plot of the score distributions for each season for the home and away teams.}
	\label{fig:box-plot-score-season-home-away}
\end{figure}

In Figure \ref{fig:box-plot-score-season-home-away}, we plot the distribution of the scores for the Home and Away teams for each of the three first seasons of the modern era. We observe that Home team distribution is shifted to higher values than the Away team distributions. Also the latter are wider for the season 2017 and 2018. Median and 75th quantile values for the Home Team distributions increase from 2017 to 2019, i.e. Home teams tend to score more.

\begin{figure}[]
	\centering
	\includegraphics[width=\textwidth]{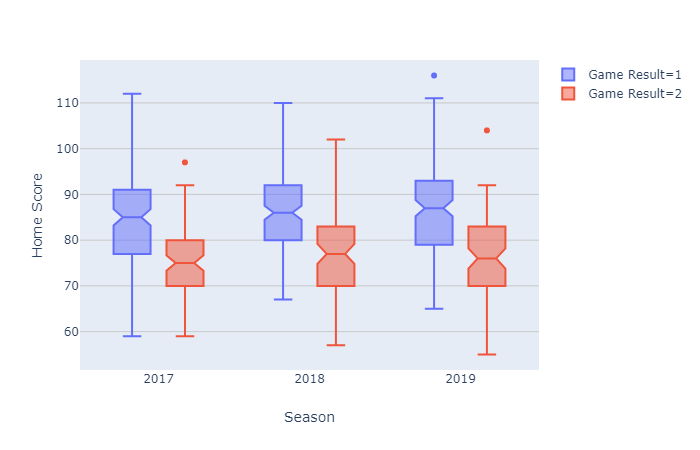}
	\caption{Box plot of the Home Team score distributions for wins (1) and losses (2).}
	\label{fig:box-plot-home-score-season-1-2}
\end{figure}

In Figure \ref{fig:box-plot-home-score-season-1-2} the Home Team Score distributions are plotted for wins (blue) and losses (red). The Win distributions are shifted to higher values in recent seasons, i.e. Home teams score more points on average each year in victories, whereas in losses the median score remains constant. Home Team victories are achieved when the Home team score is around 85 points, where in losses their median score is 76 points. 

\begin{figure}[]
	\centering
	\includegraphics[width=\textwidth]{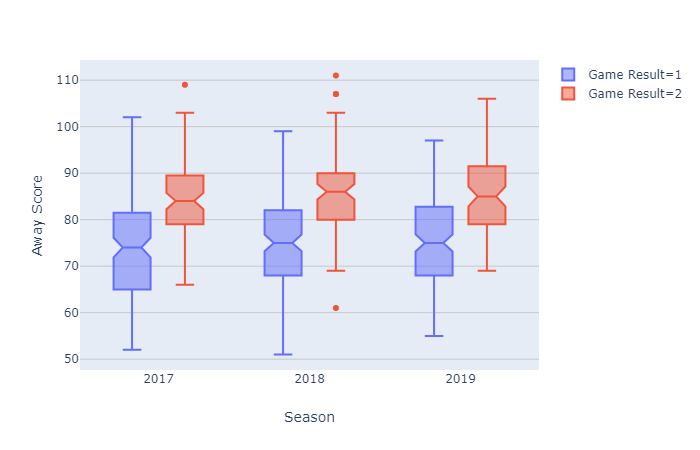}
	\caption{Box plot of the Away Team score distributions for wins (1) and losses (2).}
	\label{fig:box-plot-away-score-season-1-2}
\end{figure}

A similar plot, but for the Away Teams is shown in Figure \ref{fig:box-plot-away-score-season-1-2}. Same conclusions as before can be drawn.

\begin{figure}[]
	\centering
	\includegraphics[width=\textwidth]{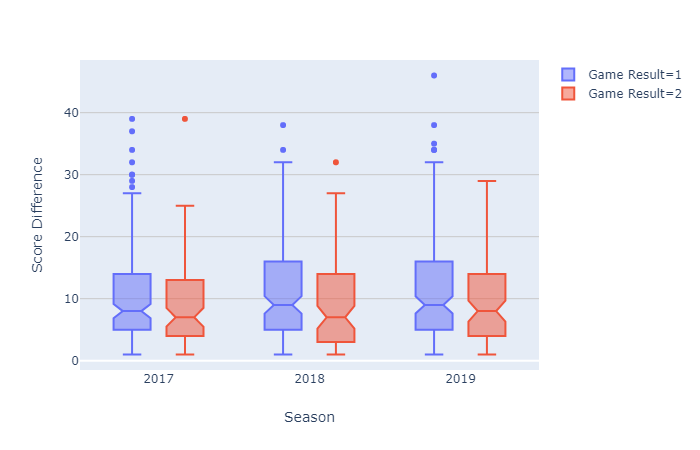}
	\caption{Box plot of the (absolute value) score difference distributions for home (1) and away (2) wins.}
	\label{fig:box-plot-diff-season-1-2}
\end{figure}

Finally, in Figure \ref{fig:box-plot-diff-season-1-2} the distribution of points difference for each season for Home and Away wins respectively. The distribution of the score difference for the Home wins remains unchanged across the years, $50\%$ (the median) of the Home wins are determined with less than 9 points difference, whereas $25\%$ (the first quartile) of the Home wins the game difference is 5 points or less. For Away wins, the distribution of differences is shifted to lower values, with $50\%$ of the Away wins end with 8 points difference or less, and $25\%$ of the Away wins end with 4 points difference or less. In Layman's term, one could say that is harder to win on the road. 
 
We summarise, a few more statics in Table \ref{tbl:summary-stats}. We notice that at least $63\%$ of the games end in Home win every year. The average Home Score has an increasing trend, same as the average Home Score of the Home wins. However, the Away scores fluctuate for seasons 2017 - 2019.

\begin{table}[]
	\centering
	\begin{tabular}{|l||p{1cm}|p{1cm}|p{1cm}|p{1cm}|p{2cm}|p{2cm}|}
		\hline
		Season & Home Wins & Away Wins & Mean Home Score & Mean Away Score & Mean Home Win Score & Mean Away Win Score \\ \hline \hline
		2017 & 152 & 88 & 80.8 & 77.5 & 78.9 & 79.6  \\ \hline
		2018 & 151 & 89 & 82.6 & 78.8 & 80.5 & 81.0  \\ \hline
		2019 & 155 & 85 & 82.8 & 78.6 & 80.6 & 80.9  \\ \hline
	\end{tabular}
	\caption{Summary statistics of the seasons 2017 - 2019.}
	\label{tbl:summary-stats}
\end{table}

It should be emphasised that these trends and patterns across the years are very preliminary, as there are only three regular seasons completed so far, and no statistically significant conclusions can be made for trends across the seasons. One conclusion that seems to arise is that Home court is an actual advantage as it has a biggest share in wins and also teams tend to score more, and hence win, when playing at Home.

\subsection{Probability of Winning}

Now, we explore the probability of a team to win when it scores more than $N$ points. Particularly, we would like to answer the question: \textit{When a team scores at least, say, 80 points, what is the probability of winning?}

\begin{figure}[]
	\centering
	\includegraphics[width=\textwidth]{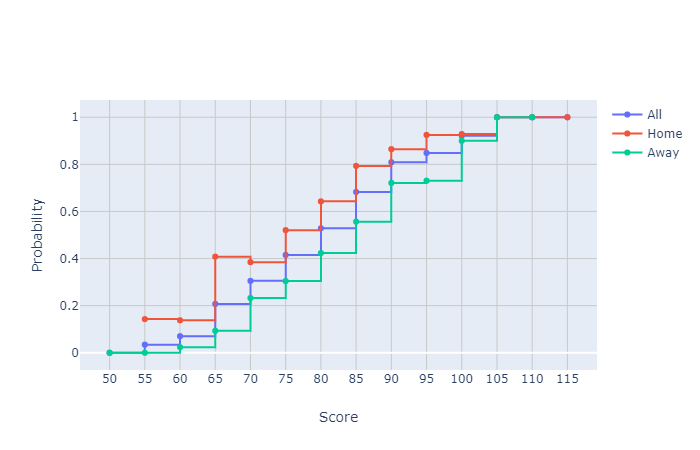}
	\caption{Probability of winning when a team scores certain points.}
	\label{fig:probability-all-home-away}
\end{figure}

The results are plotted in Figure \ref{fig:probability-all-home-away}. We plot the probability of winning for all games, the Home and the Away games respectively. It becomes evident that for the same points, a Home team is more likely to win than an Away team, which means that the defence is better when playing at home. We also observe that in Euroleague a team must score more than 80 points to have a chance better than random to win a game.

%
%
%

The conclusion of this descriptive analysis is that patterns remain unchanged across the years and no major shift in basketball style has been observed during these three years of the new format of the Euroleague. This is an important conclusion for the modelling assumptions discussed in the next section.

%% file: modelling.tex

\section{Predictive Modelling}
\label{sec:predictive-modelling}

In this section, we formulate the problem and discuss the methodology, the feature selection, calibration and testing phases. 

\subsection{Methodology}

From the machine-learning point of view, the prediction of the outcome of basketball games is a binary classification problem, one needs to classify whether a game ends as 1 (home win) or 2 (away win).

Second, we need to define the design matrix of the problem, i.e. the observations and their features. Here, we experiment with two approaches, match-level and team-level predictions, elaborated in the next subsections. 

\subsubsection{Match-level predictions}
\label{sec:match-level-predictions}

 At the match-level, the observations are the matches and the features are:

\begin{itemize}
	\item Position in the table, prior the game, of the home and away team respectively.
	
	\item The average offense of the season's matches, prior the game, of the home and away team respectively.
	
	\item The average defense of the season's matches, prior the game, of the home and away team respectively.
	
	\item The average difference between the offense and defense of the season's matches, prior the game, of the home and away team respectively.
	
	\item The form, defined as the fraction of the past five games won, of the home and away team respectively.
	
	\item The final-four flag, a binary variable indicating whether the home (resp. away) team qualified to the final-four of the \textit{previous} season.
	
\end{itemize} 	
 
At the match-level, we consider features of both the home and away teams, resulting in 12 features in total. All features quantify the teams' past performance, there is no information for the game to be modelled. These features are calculated for each season separately, as teams change significantly from one season to the next. As a result, there is no information (features) for the first game of each season, which we ignore from our analysis.

This list of features can be further expanded with more statistics, such as the average shooting percentage, assists, etc., of the home and away teams respectively. For the time-being, we focus on aforementioned simple performance statistics and try to explore their predictive power.

\subsubsection{Team-level predictions}

At team-level, the observations are teams and the machine learning model predicts the probability of a team to win. To estimate the outcome of a match, we compare the winning probabilities of the competing teams, the one with higher probability wins the match. Hence, we build a team-predictor, with the following features:
\begin{itemize}
	\item Home flag, an indicator whether the team play at home (1) or away (0).
	\item Position of the team in the table prior the game.
	\item The average offense of the season's matches of the team prior the game.
	\item The average defense of the season's matches of the team prior the game.
	\item The average difference between the offense and defense of the season's matches of the team prior the game.
	\item The form, defined as above, of the team.
	\item The final-four flag, a binary variable indicating whether the team qualified to the final-four of the \textit{previous} year.
\end{itemize}

Now, we have seven features for each observation (i.e. team), but we also doubled the number of observations compared to the match-level approach. The machine-learning model predicts the probability of a team to win, and these probabilities are combined to predict the outcome of a game.

As mentioned before, these features can be expanded to include percentage shootings, steals, etc. We leave this for future work.

For the remaining of the analysis, we split the data into two sub-sets, the 2017, 2018 seasons for hyper-parameter calibration, feature selection and training, and the final season 2019 for testing. Also, all features are normalised to the interval $[0, 1]$, so that biases due to magnitude are removed.

\subsubsection{Classifiers}

The binary classification problem is a well-studied problem in the machine-learning literature and numerous algorithms have been proposed and developed \cite{Hastie2009elements}. Here, we experiment with the following list of the off-the-shelves algorithms that are available in the \textit{scikit-learn} Python library \footnote{\url{https://scikit-learn.org/stable/}} \cite{scikit-learn}. The reason we use these algorithms is two-fold; (i) we do not want to re-invent the wheel, we use well-established algorithms (ii) the \textit{scikit-learn} library has been developed and maintained at high-standards by the machine-learning community, constituting an established tool for the field.

The classifiers are:
\begin{itemize}
	\item Logistic Regression (LR)
	\item Support Vector Machines (SVM) with linear and \textit{RBF} kernels
	\item Decision Tree (DT)
	\item Random Forest (RF)
	\item Naive Bayes (NB)
	\item Gradient Boosting (GB)
	\item $k$-Nearest Neighbours ($k$NN)
	\item Discriminant Analysis (DA)
	\item AdaBoost (Ada)
\end{itemize}

\subsection{Calibration}
\label{sec:calibration}

The above classifiers have hyper-parameters, which we tune using 5-fold cross-validation and grid-search on the training set. Although some algorithms have more than one hyper-parameters, in the majority of the classifiers, we limit ourselves to tuning only a few parameters in order to speed the computations. These hyper-parameters are:
\begin{itemize}
	\item The regularisation strength for LR and SVM-linear
	\item Number of estimators for RF, GB and Ada
	\item Number of neighbours for $k$NN
	\item The regularisation strength and the width of the RBF kernel for the SVM-rbf
	\item Number of estimators and learning rate for Ada, abbreviated as Ada2~\footnote{We experiment with two versions of the Ada classifiers.}
\end{itemize}
All other hyper-parameters are set to their default values.

\begin{figure}[]
	\centering
	\includegraphics[width=\textwidth]{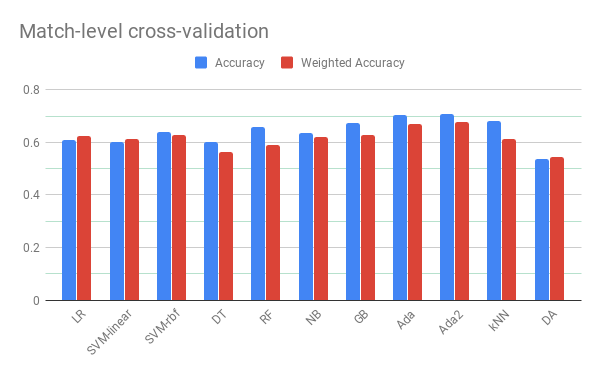}
	\caption{Accuracy (blue) and weighted accuracy (red) scores of 5-fold cross-validation at the match-level classification.}
	\label{fig:match-level-cross-validation}
\end{figure}

We record both the accuracy and the weighted accuracy across the folds. For each value of the hyper-parameter search, we find the mean score across the folds and report its maximum value. In Figure \ref{fig:match-level-cross-validation}, we plot the scores of the 5-fold cross validation process for the aforementioned classifiers at the match-level classification. We observe that the Ada and Ada2 classifiers outperform all others for all scores, with gradient boosting coming third.

\begin{figure}[]
	\centering
	\includegraphics[width=\textwidth]{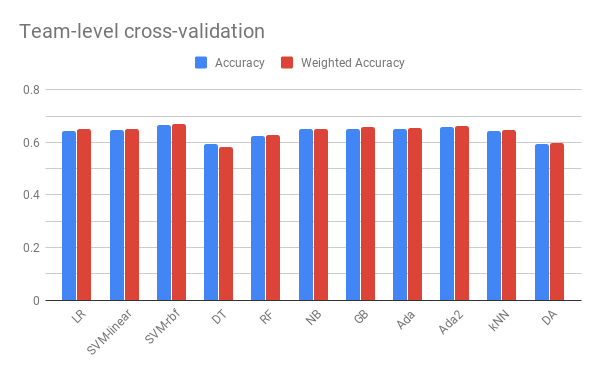}
	\caption{Accuracy (blue) and weighted accuracy (red) scores of 5-fold cross-validation at the team-level classification.}
	\label{fig:team-level-cross-validation}
\end{figure}

In Figure \ref{fig:team-level-cross-validation}, we plot the scores of the 5-fold cross validation for the classifiers of the team-level analysis. We observe that the team-level analysis is under-performing the match-level models for almost all classifiers.

We conclude that the Ada and Ada2 classifiers at the match-level are the best-performing models, having accuracy score 0.705 and 0.708 respectively, and we select them for further analysis.

\subsection{Feature Selection}
\label{sec:feature-selection}

Feature selection is an important step in machine-learning methodoloty. There are several methods for selecting the most informative features; filter, embedded, wrapper and feature transformation (e.g. Principal Component Analysis (PCA)) methods, see \cite{Guyon:2003,Verleysen:2005} for further details. Here, we explore the aforementioned methods and attempt to understand the most informative features. 

\subsubsection{Filter Methods}

Filter methods are agnostic to the algorithm and using statistical tests they aim to determine the features with the highest statistical dependency between the target and feature variables \cite{Guyon:2003,Verleysen:2005}. For classification, such methods are: (i) the ANOVA F-test, (ii) the mutual information (MI) and (iii) chi-square (Chi2) test.

\begin{figure}[]
	\centering
	\includegraphics[scale=0.5]{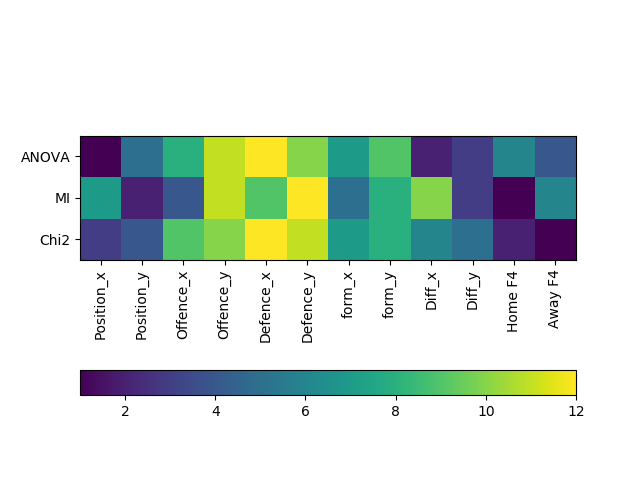}
	\caption{Ranking (1 most, 12 lest informative) of features according to the ANOVA, mutual information (``MI'') and Chi-square (``Chi2'') filter methods. The ``x'' and ``y'' suffixes refer to the features of the Home and Away team respectively.} 
	\label{fig:filter-ranking}
\end{figure}

We use the match-level features and apply the three filter methods. We rank the features from the most informative to the least informative for the three methods. We plot the results in Figure \ref{fig:filter-ranking}. The darker blue a feature is, the most informative it is for the corresponding method. On the other extreme, least informative features are yellowish. We observe some common patterns, the position and F4 features are very informative, whereas the Offence of the Away team and the Defence features are the least informative for all methods. 

\begin{figure}
	\centering
	\begin{subfigure}[b]{0.3\textwidth}
		\includegraphics[width=\textwidth]{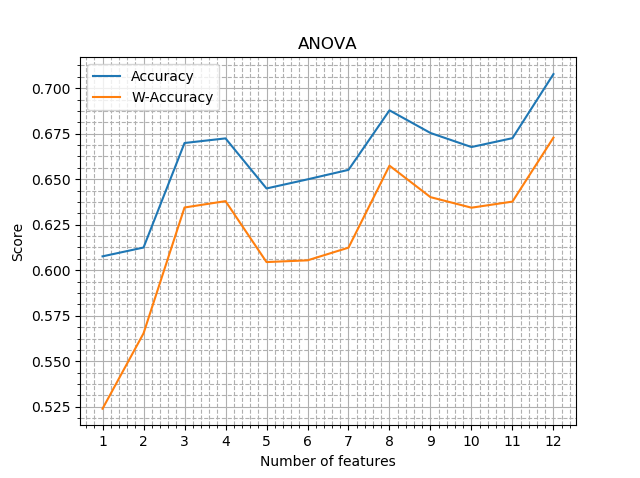}
		\caption{ANOVA}
		\label{fig:filter-anova-ada}
	\end{subfigure}
	~
	\begin{subfigure}[b]{0.3\textwidth}
		\includegraphics[width=\textwidth]{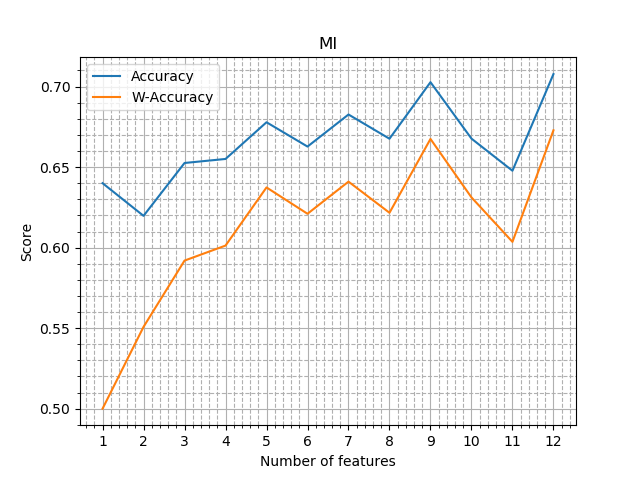}
		\caption{Mutual Information}
		\label{fig:filter-mi-ada}
	\end{subfigure}
	~
	\begin{subfigure}[b]{0.3\textwidth}
		\includegraphics[width=\textwidth]{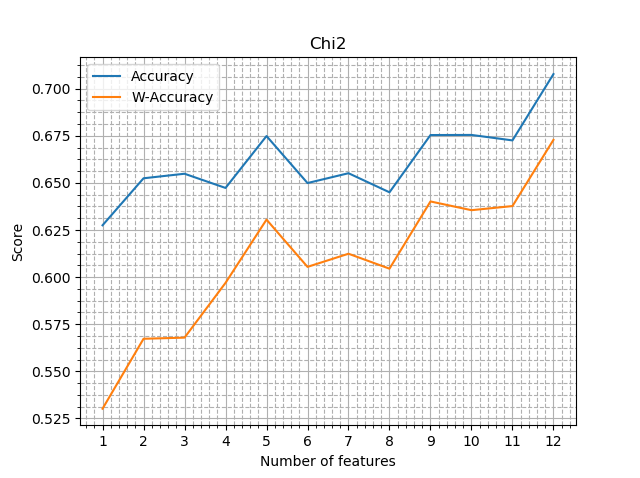}
		\caption{Chi-square}
		\label{fig:filter-chi2-ada}
	\end{subfigure}
	\caption{Performance socres for increasing number of features, starting with the most informative features for each method.}
	\label{fig:filter-components}
\end{figure}

If some features are less informative, we assess the performance of the model with increasing number of features, from the most informative according to each filter method, adding the less informative ones, incrementally. If some features have negative impact to the performance of the models, we expect the accuracy to peak at some features and then decrease for the rest. However, we do observe, see Figure \ref{fig:filter-components}, that the maximum performance scores (accuracy and weighted accuracy) are achieved when all 12 features are included in the model. Hence, although some features are less informative, they all contribute positively to the model's performance.

As no subset of features can exceed the performance of "all features" model, we discard filter methods as a potential method for feature reduction/selection.

\subsubsection{PCA}

Principal Component Analysis (PCA) is a dimensional reduction of the feature space method, see \cite{Hastie2009elements} for further details. We run the PCA for increasing number of components, from 1 to the maximum number of features and assess the number of components using the Ada model. In addition, at each value, grid-search finds the optimal hyper-parameter (the number of estimators) and optimal model is assessed \footnote{The reason we re-run the grid-search is that the hyper-parameter(s) found in the previous section is optimal to the original feature-set. PCA transforms the features and hence a new hyper-parameter might be optimal.}.
 
\begin{figure}[]
	\centering
	\includegraphics[scale=0.7]{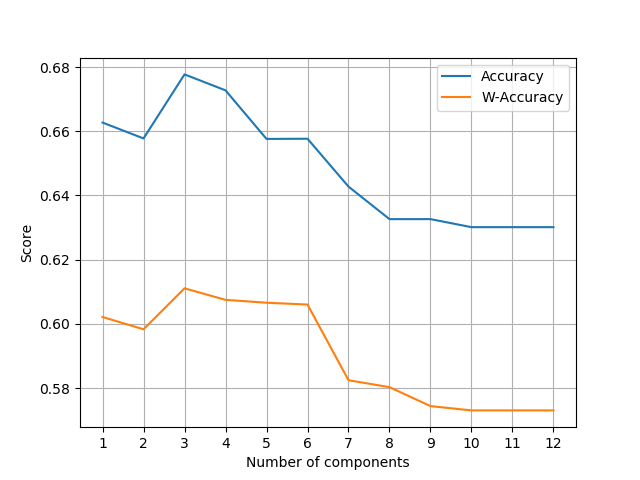}
	\caption{Accuracy and weighted accuracy for increasing number of principal components.}
	\label{fig:pca-ada}
\end{figure}

The best accuracy and weighted-accuracy scores for different number of components are plotted in Figure \ref{fig:pca-ada}. First, we observe that the model's performance peaks at three components, which capture most of the variance. As more components are added, noise is added and the performance decreases. Second, even at peak performance  the accuracy and weighted accuracy do not exceed those of the Ada model with the originals features. Hence, we discard PCA as a potential method for feature reduction/selection.

\subsubsection{Wrapper Methods}
\label{sec:wrapper-methods}

Wrapper methods scan combinations of features and assess their performance using a classification algorithm and select the best feature (sub-)set that maximises the performance score (in a cross-validation fashion). Hence, they depend on the classification method. 

Since the number of features in our study is relatively small ($12$), we are able to scan all possible combinations of features subsets ($4,095$ in total). As we cannot find the optimal hyper-parameters for each subset of features (the search space becomes huge), we proceed in a two-step iterative process. First, we use the best pair of hyper-parameters of the ``Ada2'' algorithm\footnote{We remind the reader, that we focus on the ``Ada2'' algorithm, as it is shown in the previous section that it is the best-performing algorithm for this task.} to scan all the combinations of features. Then, we select the combination of features which outperforms the others. In the second step, we re-train the algorithm to find a new pair of hyper-parameters that might be better for the selected feature subset.

\begin{figure}
	\centering
	\begin{subfigure}[b]{0.45\textwidth}
		\includegraphics[width=\textwidth]{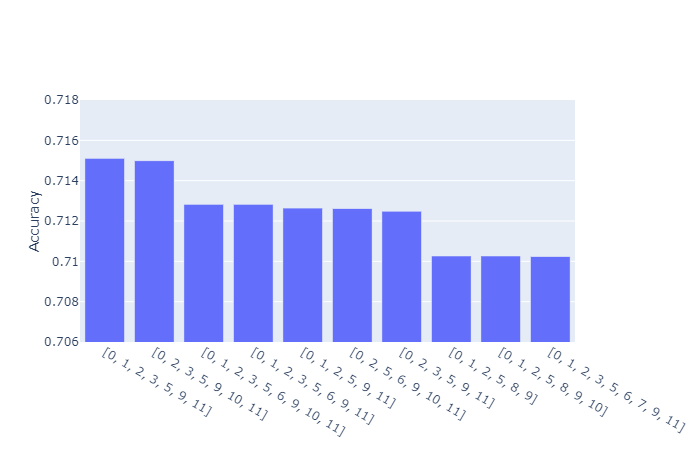}
		\caption{Accuracy}
		\label{fig:wrapper-accuracy}
	\end{subfigure}
	~
	\begin{subfigure}[b]{0.45\textwidth}
		\includegraphics[width=\textwidth]{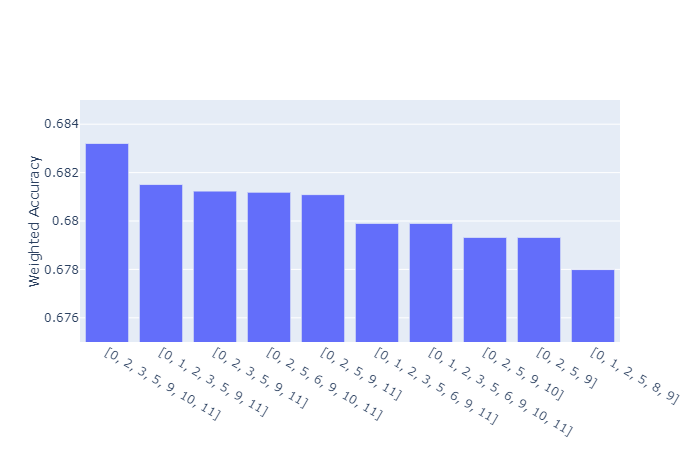}
		\caption{Weighted Accuracy}
		\label{fig:wrapper-w-accuracy}
	\end{subfigure}
	\caption{Performance scores of the top-10 sub-sets of features via the wrapper method of feature selection. The x-axis indicates the index (zero-based) of the feature in the following list ['Position Home', 'Position Away', 'Offence Home', 'Offence Away', 'Defence Home', 'Defence Away', 'Form Home', 'Form Away',	'Difference Home', 'Difference Away', 'Home F4', 'Away F4'], see also Section \ref{sec:match-level-predictions}}
	\label{fig:wrapper-method}
\end{figure}

In first step, we sort the different combinations of features by their performance and plot the accuracy and weighted accuracy of the top ten combination in Figure \ref{fig:wrapper-method}. We observe that the two sub-sets of features are in the top two feature sub-set both in accuracy and weighted accuracy. We select these two feature sub-sets for further evaluation, hyper-parameter tuning, final training and testing. These feature tests are:
\begin{itemize}
	\item Model 1: Position Home, Position Away, Offence Home, Offence Away, Defence Away, Difference Away, F4 Away 
	\item Model 2: Position Home, Offence Home, Offence Away, Defence Away, Difference Away, F4 Home, F4 Away.
\end{itemize}

Having selected these two sets of features, we proceed to the step two of the iterative process and re-calibrate their hyper-parameters. At the end of the two-step iterative process, we identify the following subset of features with their optimal hyper-parameters:
\begin{itemize}
	\item Model 1: Ada2 with 141 estimators and 0.7 learning rate achieving 0.7498 accuracy and 0.7222 weighted accuracy.
	\item Model 2: Ada2 with 115 estimators and 0.7 learning rate, achieving 0.7375 accuracy and 0.7066 weighted accuracy.
\end{itemize}

We conclude that the wrapper method results in the best performing features and hence model, achieving an accuracy of 75\% via 5-fold cross-validation. We select these models to proceed to the validation phase.

\subsection{Validation}
\label{sec:validation}

In this section, we use the models (i.e. features and hyper-parameters) selected in the previous section for validation in the season 2018-2019. For comparison, we also include the Ada2 model with all features (without feature selection) as ``Model 3''.
At each game round, we train the models given all the available data before that round. For example, to predict the results of round 8, we use all data available from seasons 2017, 2018 and the rounds 2 to 7 from season 2019. We ignore round 1 in all seasons, as no information (i.e. features) is still available for that round. 
\footnote{We comment on the ``cold-start'' problem in the Discussion section.} 
Then we predict the results of the 8 games of that round, we store the results and proceed to the next round.

We benchmark the models against three simple models. These are:
\begin{itemize}
	\item Benchmark 1: Home team always wins. This is the majority model in machine learning terms.
	\item Benchmark 2: The Final-4 teams of the previous year always win, if there are no Final-4 teams playing against each other, or both teams are Final-4 teams, the home team always wins.
	\item Benchmark 3: The team higher in the standings, at that moment in the season, wins.
\end{itemize}

\begin{table}[]
	\centering
	\begin{tabular}{l||r|r}
		& Accuracy & Weighted Accuracy \\
		\hline
		Model 1     & 0.6595   & 0.6096            \\
		Model 2     & 0.6595   & 0.6211            \\
		Model 3     & 0.6681   & 0.6306            \\
		Benchmark 1 & 0.6509   & 0.5000            \\
		Benchmark 2 & 0.7198   & 0.6446            \\
		Benchmark 3 & 0.6810   & 0.7035           
	\end{tabular}
	\caption{Accuracy and weighted accuracy of the three models and three benchmarks.}
	\label{tbl:validation-scores-benchmarking}
\end{table}

From Table \ref{tbl:validation-scores-benchmarking} we observe the following. First, the machine-learning models (marginally) outperform the majority benchmark. However, they fail to outperform the more sophisticated benchmarks 2 and 3. We also observe that the models 1 and 2 resulting from feature selection perform worse than model 3 (all features are present). This implies that the models 1 and 2 have been over-fitted and no feature selection was required at this stage as the number of features is still relatively small. This is also in agreement with the filter methods feature selection. The performance of benchmark 2 is a special case, as the season 2018-2019 the teams that qualified to the Final-4 of the previous season (2017-2018) continued their good basketball performance. However, historically, this is not always the case and it is expected that this benchmark not to be robust for other seasons.

\begin{figure}
	\centering
	\includegraphics[width=\textwidth]{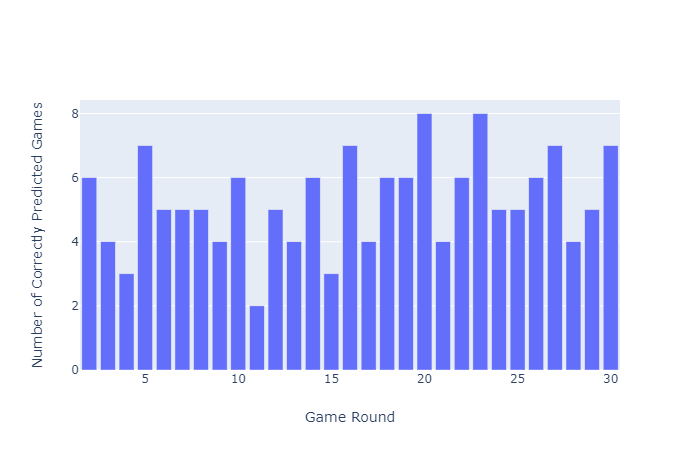}
	\caption{Number of correct games per round}
	\label{fig:corrects-per-round}
\end{figure}

\begin{figure}
	\centering
	\includegraphics[width=\textwidth]{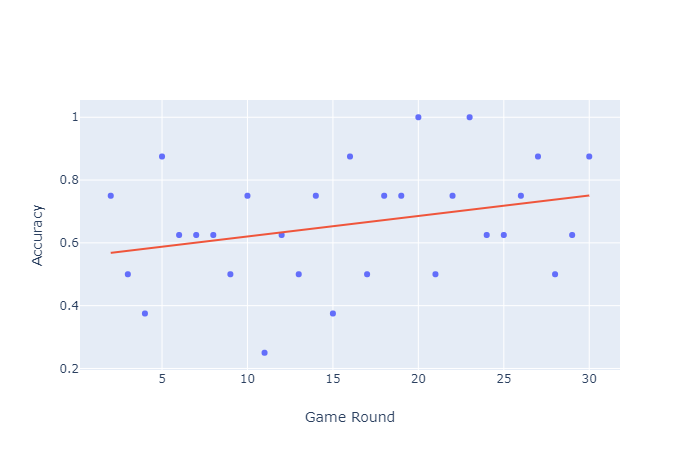}
	\caption{Accuracy and the trend of accuracy.}
	\label{fig:accuracy-per-round}
\end{figure}

We focus on ``Model 2'' and plot the number of correct games in every round of the season 2018-2019 in Figure \ref{fig:corrects-per-round}, and the model's accuracy per round in \ref{fig:accuracy-per-round}. From these Figures, we observe that the model finds perfect accuracy in rounds 20 and 23, whereas it achieves 7/8 in four other rounds. With the exception of 4 rounds the model always predicts at least half of the games correctly. We manually inspected round 11, where the model has its lower accuracy, and there were 4 games that a basketball expert would consider ``unexpected'', a factor that might contribute to the poor performance of the model in that round.

In Figure \ref{fig:accuracy-per-round} we also plot the trend of the accuracy, which is positive, and indicates that as the seasons progresses, the teams become more stable and the model more accurate.

\section{The Wisdom of (Basketball) Crowds}
The author is a member of a basketball forum on a social networking platform. On this forum, a basketball predictions championship is organised for fun. In this championship, the members give their predictions for the Euroleague games in the upcoming round. They get 1 point if the prediction is correct, 0 otherwise. At the end of the season the players that come in the top eight of the table qualify to the play-offs phase. 

Here, we collect the players' predictions. For each game in each round, we assign a result, which is the result of the majority prediction of the players~\footnote{Round 26 is omitted, due to a technical issue on the forum, results were not collected and the round was invalid.}. The championship started with about 60 players and finished with 41 players still active, as many dropped out during the course of the season.

We update the Table \ref{tbl:validation-scores-woc} to include the majority vote of the members of the forum (but excluding the 26th round in the accuracy due to the technical issue - first round is excluded in all performance scores.). 

We observe that the collective predictions of the members of the forum have a much better accuracy (and weighted accuracy) than any model in this study (machine-learning or benchmark). This is a known phenomenon in collective behaviour, termed the ``wisdom of the crowds'' \cite{Surowiecki2005woc}, observed in social networks and rumour detection \cite{Giasemidis:2016dtv} among others. For this study, we adapt the term and refer to it as ``the wisdom of the \textit{basketball} crowds''.

\begin{table}[]
	\centering
	\begin{tabular}{l||r|r}
		& Accuracy & Weighted Accuracy \\
		\hline
		Model 1     & 0.6607   & 0.6113            \\
		Model 2     & 0.6563   & 0.6199            \\
		Model 3     & 0.6696   & 0.6331            \\
		Benchmark 1 & 0.6518   & 0.5000            \\
		Benchmark 2 & 0.7188   & 0.6449            \\
		Benchmark 3 & 0.6786   & 0.7027            \\
		Majority Vote of the Members & 0.7321   & 0.6811 
	\end{tabular}
	\caption{Accuracy and weighted accuracy of the three models, three benchmarks and the forum results.}
	\label{tbl:validation-scores-woc}
\end{table}

%% file: conclusions.tex

\section{Summary and Conclusions}
\label{sec:conclusions}

In this article we focused on Euroleague basketball games in the modern format era. We first explored their descriptive characteristics of the last three seasons and highlighted the importance of home advantage. Also, the game characteristics remain stable across the three seasons, allowing us to use data across the seasons for modelling purposes.

We then introduced a framework for machine-learning modelling of the games' outcomes. This is a binary classification task in machine learning terms. We compared the nine most commonly used algorithms for classification. The AdaBoost algorithm outperforms the others in this classification task. 
Two classification methodologies were also compared, match-level and team-level classification, with the former being more accurate.
Feature selection and hyper-parameter tuning are very essential steps in the machine learning pipeline, and previous studies had ignored either or both of these steps. Wrapper methods proved to outperform filter methods and PCA. 
Also, proper validation of the models on a hold-out set and their benchmarking had been omitted in some of the previous studies (most of them validated their models using $k$-fold cross validation). 

Our model achieves 75\% accuracy using 5-fold cross validation. However, this is not the complete picture. The model was properly validated on a hold-out set, the season 2018--2019, achieving accuracy 66.8\%. This indicates that either the models were over-fitted during $k$-fold cross validation or the season 2018--2019 is exceptionally difficult to predict. However, the success of simple benchmarks shows that over-fitting is more likely to be the case. Over-fitting during the training phase (including feature selection and hyper-parameter tuning) is always a risk in machine learning. One lesson taken is that $k$-fold cross validation is not always an informative and conclusive way to validate a model.

We also compared our results to the collective predictions of a group of basketball enthusiasts. The collective predictive power of that group outperforms any model (machine learning or benchmark) in this study. We are intrigued by this result. We believe that any machine learning model to be considered successful should outperform the human predicting capabilities (either individual or collective). For this reason, we set the accuracy of 73\% as the baseline for accuracy of Euroleague basketball games. This is also in agreement with \cite{zimmermann2013predicting}, in which the authors found that all machine learning models had a predictive accuracy bounded by 74\% (althouth this was for NBA games). Future models for basketball predictions should aim to break this threshold.

\subsection{What is next?}
There are many potential directions and improvements of this study. First, we aim to repeat this analysis with more team features, such as field goals percentages, field goals made, three pointers, free throws, blocks, steals (normalised either per game or per possesion), etc. Additional features from  advanced statistics could be included, such as effective field goals, efficiencies, etc. The feature space then becomes larger and feature reduction techniques might prove meaningful.

Another direction for improvement is the inclusion of players' information and features. We discussed that the study  in \cite{Oursky2019mla} achieves 80\% accuracy in NBA game predictions by including players' features in their classifier. 

Finally, as the seasons progress higher volumes of data will become available. Although more data does not necessarily lead to more accurate models, we will be able to conduct more analysis and potentially avoid over-fitting. In many studies we reviewed in the Introduction, the accuracy of the models vary greatly from one season to the next. More data will provide us with more robust results. Additionally, we will be able to exploit methods such as neural networks which proved successful in \cite{Oursky2019mla}.

This study is neither a final solution nor the end-story. In contrast, it lays the methodology and best-practices for further exploring the question of predictions of game outcomes. We will continue re-train and re-evaluate our models with the 73\% accuracy threshold (the accuracy of the wisdom basketball crowds) in mind.